\documentclass[10pt,twocolumn,letterpaper]{article}

\usepackage{cvpr}
\usepackage{times}
\usepackage{epsfig}
\usepackage{graphicx}
\usepackage{amsmath}
\usepackage{commath}
\usepackage{amssymb}
\usepackage{enumerate,booktabs}
\usepackage{subfigure,color}
\usepackage[numbers,sort&compress]{natbib}
\usepackage{multirow}
\usepackage{pifont}

\usepackage[pagebackref=true,breaklinks=true,letterpaper=true,colorlinks,bookmarks=false]{hyperref}

\cvprfinalcopy % *** Uncomment this line for the final submission
\def\cvprPaperID{710} % *** Enter the CVPR Paper ID here

\ifcvprfinal\fi
\begin{document}
\def\mA{\mathcal{A}}
\def\mB{\mathcal{B}}
\def\mC{\mathcal{C}}
\def\mD{\mathcal{D}}
\def\mE{\mathcal{E}}
\def\mF{\mathcal{F}}
\def\mG{\mathcal{G}}
\def\mH{\mathcal{H}}
\def\mI{\mathcal{I}}
\def\mJ{\mathcal{J}}
\def\mK{\mathcal{K}}
\def\mL{\mathcal{L}}
\def\mM{\mathcal{M}}
\def\mN{\mathcal{N}}
\def\mO{\mathcal{O}}
\def\mP{\mathcal{P}}
\def\mQ{\mathcal{Q}}
\def\mR{\mathcal{R}}
\def\mS{\mathcal{S}}
\def\mT{\mathcal{T}}
\def\mU{\mathcal{U}}
\def\mV{\mathcal{V}}
\def\mW{\mathcal{W}}
\def\mX{\mathcal{X}}
\def\mY{\mathcal{Y}}
\def\mZ{\mathcal{Z}}

\def\1n{\mathbf{1}_n}
\def\0{\mathbf{0}}
\def\1{\mathbf{1}}

\def\A{{\bf A}}
\def\B{{\bf B}}
\def\C{{\bf C}}
\def\D{{\bf D}}
\def\E{{\bf E}}
\def\F{{\bf F}}
\def\G{{\bf G}}
\def\H{{\bf H}}
\def\I{{\bf I}}
\def\J{{\bf J}}
\def\K{{\bf K}}
\def\L{{\bf L}}
\def\M{{\bf M}}
\def\N{{\bf N}}
\def\O{{\bf O}}
\def\P{{\bf P}}
\def\Q{{\bf Q}}
\def\R{{\bf R}}
\def\S{{\bf S}}
\def\T{{\bf T}}
\def\U{{\bf U}}
\def\V{{\bf V}}
\def\W{{\bf W}}
\def\X{{\bf X}}
\def\Y{{\bf Y}}
\def\Z{{\bf Z}}

\def\a{{\bf a}}
\def\b{{\bf b}}
\def\c{{\bf c}}
\def\d{{\bf d}}
\def\e{{\bf e}}
\def\f{{\bf f}}
\def\g{{\bf g}}
\def\h{{\bf h}}
\def\i{{\bf i}}
\def\j{{\bf j}}
\def\k{{\bf k}}
\def\l{{\bf l}}
\def\m{{\bf m}}
\def\n{{\bf n}}
\def\o{{\bf o}}
\def\p{{\bf p}}
\def\q{{\bf q}}
\def\r{{\bf r}}
\def\s{{\bf s}}
\def\t{{\bf t}}
\def\u{{\bf u}}
\def\v{{\bf v}}
\def\w{{\bf w}}
\def\x{{\bf x}}
\def\y{{\bf y}}
\def\z{{\bf z}}

\def\balpha{\mbox{\boldmath{$\alpha$}}}
\def\bbeta{\mbox{\boldmath{$\beta$}}}
\def\bdelta{\mbox{\boldmath{$\delta$}}}
\def\bgamma{\mbox{\boldmath{$\gamma$}}}
\def\blambda{\mbox{\boldmath{$\lambda$}}}
\def\bsigma{\mbox{\boldmath{$\sigma$}}}
\def\btheta{\mbox{\boldmath{$\theta$}}}
\def\bomega{\mbox{\boldmath{$\omega$}}}
\def\bxi{\mbox{\boldmath{$\xi$}}}
\def\bnu{\mbox{\boldmath{$\nu$}}}                                  
\def\bphi{\mbox{\boldmath{$\phi$}}}
\def\bmu{\mbox{\boldmath{$\mu$}}}

\def\bDelta{\mbox{\boldmath{$\Delta$}}}
\def\bOmega{\mbox{\boldmath{$\Omega$}}}
\def\bPhi{\mbox{\boldmath{$\Phi$}}}
\def\bLambda{\mbox{\boldmath{$\Lambda$}}}
\def\bSigma{\mbox{\boldmath{$\Sigma$}}}
\def\bGamma{\mbox{\boldmath{$\Gamma$}}}

\newcommand{\myminimum}[1]{\mathop{\textrm{minimum}}_{#1}}
\newcommand{\mymaximum}[1]{\mathop{\textrm{maximum}}_{#1}}    
\newcommand{\mymin}[1]{\mathop{\textrm{minimize}}_{#1}}
\newcommand{\mymax}[1]{\mathop{\textrm{maximize}}_{#1}}
\newcommand{\mymins}[1]{\mathop{\textrm{min.}}_{#1}}
\newcommand{\mymaxs}[1]{\mathop{\textrm{max.}}_{#1}}  
\newcommand{\myargmin}[1]{\mathop{\textrm{argmin}}_{#1}} 
\newcommand{\myargmax}[1]{\mathop{\textrm{argmax}}_{#1}} 
\newcommand{\myst}{\textrm{s.t. }}

\newcommand{\denselist}{\itemsep -1pt}
\newcommand{\sparselist}{\itemsep 1pt}

\definecolor{pink}{rgb}{0.9,0.5,0.5}
\definecolor{purple}{rgb}{0.5, 0.4, 0.8}   
\definecolor{gray}{rgb}{0.3, 0.3, 0.3}
\definecolor{mygreen}{rgb}{0.2, 0.6, 0.2}

\newcommand{\cyan}[1]{\textcolor{cyan}{#1}}
\newcommand{\red}[1]{\textcolor{red}{#1}}  
\newcommand{\blue}[1]{\textcolor{blue}{#1}}
\newcommand{\magenta}[1]{\textcolor{magenta}{#1}}
\newcommand{\pink}[1]{\textcolor{pink}{#1}}
\newcommand{\green}[1]{\textcolor{green}{#1}} 
\newcommand{\gray}[1]{\textcolor{gray}{#1}}    
\newcommand{\mygreen}[1]{\textcolor{mygreen}{#1}}    
\newcommand{\purple}[1]{\textcolor{purple}{#1}}       

\definecolor{greena}{rgb}{0.4, 0.5, 0.1}
\newcommand{\greena}[1]{\textcolor{greena}{#1}}

\definecolor{bluea}{rgb}{0, 0.4, 0.6}
\newcommand{\bluea}[1]{\textcolor{bluea}{#1}}
\definecolor{reda}{rgb}{0.6, 0.2, 0.1}
\newcommand{\reda}[1]{\textcolor{reda}{#1}}

\def\changemargin#1#2{\list{}{\rightmargin#2\leftmargin#1}\item[]}
\let\endchangemargin=\endlist
                                               
\newcommand{\cm}[1]{}

\newcommand{\mtodo}[1]{{\color{red}$\blacksquare$\textbf{[TODO: #1]}}}
\newcommand{\myheading}[1]{\vspace{1ex}\noindent \textbf{#1}}
\newcommand{\htimesw}[2]{\mbox{$#1$$\times$$#2$}}

% The following are useful for creating homework or exams

\newif\ifshowsolution
%\showsolutionfalse
\showsolutiontrue

\ifshowsolution  
\newcommand{\Comment}[1]{\paragraph{\bf $\bigstar $ COMMENT:} {\sf #1} \bigskip}
\newcommand{\Solution}[2]{\paragraph{\bf $\bigstar $ SOLUTION:} {\sf #2} }
\newcommand{\Mistake}[2]{\paragraph{\bf $\blacksquare$ COMMON MISTAKE #1:} {\sf #2} \bigskip}
\else
\newcommand{\Solution}[2]{\vspace{#1}}
\fi

\newcommand{\truefalse}{
\begin{enumerate}
	\item True
	\item False
\end{enumerate}
}

\newcommand{\yesno}{
\begin{enumerate}
	\item Yes
	\item No
\end{enumerate}
}

\newcommand{\haibin}[1]{\textcolor{blue}{\bf [Haibin: #1]}}
\newcommand{\yuqian}[1]{\textcolor{green}{\bf [Yuqian: #1]}}

\newcommand{\xmark}{\ding{55}}
\newcommand{\cmark}{\ding{51}}

%%%%%%%%% TITLE
\title{GIF2Video: Color Dequantization and Temporal Interpolation of GIF images}

\author{
Yang Wang$^1$, ~Haibin Huang$^{2\dagger}$, ~Chuan Wang$^2$, ~Tong He$^3$, ~Jue Wang$^2$, ~Minh Hoai$^1$ \\
$^1$Stony Brook University,~~ $^2$Megvii Research USA,~~ $^3$UCLA,~~ $^\dagger$Corresponding Author
}

\maketitle
%\thispagestyle{empty}

%%%%%%%%% ABSTRACT
\begin{abstract}

%Graphics Interchange Format (GIF) is a highly portable image format, and GIF images are ubiquitous  very popular on the Internet, especially among social network users these days. However, as a compromise for portability, GIFs  usually contain undesirable artifacts such as flat color regions, false contours, color shift, and dotted patterns due the conversion from video. In this paper, 

Graphics Interchange Format (GIF) is a highly portable graphics format that is ubiquitous on the Internet. Despite their small sizes, GIF images often contain undesirable visual artifacts such as flat color regions, false contours, color shift, and dotted patterns. In this paper, we propose GIF2Video, the first learning-based method for enhancing the visual quality of GIFs in the wild. We focus on the challenging task of GIF restoration by recovering information lost in the three steps of GIF creation: frame sampling, color quantization, and color dithering. We first propose a novel CNN architecture for color dequantization. It is built upon a compositional architecture for multi-step color correction, with a comprehensive loss function designed to handle large quantization errors. We then adapt the SuperSlomo network for temporal interpolation of GIF frames. We introduce two large datasets, namely GIF-Faces and GIF-Moments, for both training and evaluation. Experimental results show that our method can significantly improve the visual quality of GIFs, and outperforms direct baseline and state-of-the-art approaches. 

%To reverse the process of color quantization and dithering, we present a novel CNN architecture for color dequantization. The proposed network is based on a compositional architecture for multi-step color correction to handle large quantization error. The proposed network has the color quantization process embedded to guide its training and inference. We also propose a comprehensive loss function for training the network. To reverse the process of frame sampling, we adapt the SuperSlomo network for temporal interpolation of GIF frames. 
%The proposed GIF2Video approach combines both the color dequantization network and the frame interpolation network.  
%including FCDR (False Contour Detection \& Removal) and Pix2Pix.

\end{abstract}

\section{Introduction}

\begin{figure}[t]
\begin{center}
\includegraphics[width=\linewidth]{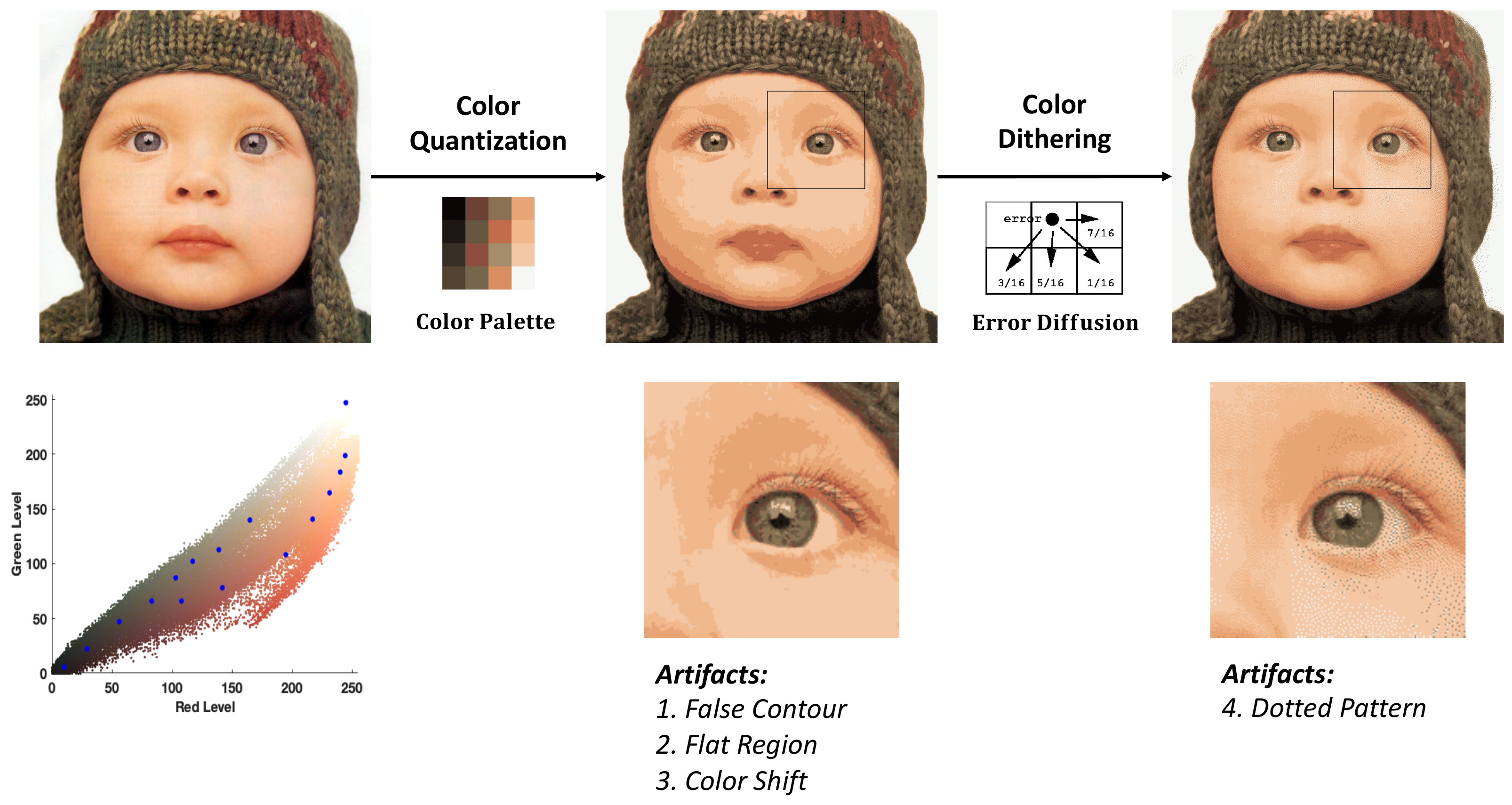}
\end{center}
\vskip -0.1in
   \caption{{\bf Color quantization and color dithering.} Two major steps in the creation of a GIF image. These are lossy compression processes that result in undesirable visual artifacts. Our approach is able to remove these artifacts and produce a much more natural image.}
\label{fig: gif_generation}
\end{figure}

GIFs~\cite{giflink} are everywhere, being created and consumed by millions of Internet users every day on the Internet. The widespread of GIFs can be attributed to its high portability and small file sizes. 
However, due to heavy quantization in the creation process, 
GIFs often have much worse visual quality than their original source videos. Creating an animated GIF from a video involves three major steps: frame sampling, color quantization, and optional color dithering. 
%Spatial downsampling is not considered because we observe the image size of the GIFs on the internet is usually sufficiently large. The visual quality of GIFs is mainly affected by other factors aforementioned. 
Frame sampling introduces jerky motion, while color quantization and color dithering create flat color regions, false contours, color shift, and dotted patterns, as shown in Fig.~\ref{fig: gif_generation}.

In this paper, we propose GIF2Video, the first learning-based method for enhancing the visual quality of GIFs. Our algorithm consists of two components. First, it performs color dequantization for each frame of the animated gif sequence, removing the artifacts introduced by both color quantization and color dithering. Second, it increases the temporal resolution of the image sequence by using a modified SuperSlomo~\cite{superslomo} network for temporal interpolation.

The main effort of this work is to develop a method for color dequantization, i.e., removing the visual artifacts introduced by heavy color quantization. Color quantization is a lossy compression process that remaps original pixel colors to a limited set of entries in a small color palette. This process introduces quantization artifacts, similar to those observed when the bit depth of an image is reduced. For example, when the image bit depth is reduced from $48$-bit to $24$-bit, the size of the color palette shrinks from $2.8 \times 10^{14}$ colors to $1.7\times 10^7$ colors, leading to a small amount of artifacts. The color quantization process for GIF, however, is far more aggressive with a typical palette of 256 distinct colors or less.  Our task is to perform dequantization from a tiny color palette (e.g., 256 or 32 colors), and it is much more challenging than traditional bit depth enhancement~\cite{wan2016bitMAP, liu2017bitCNN, huang2018fcdr}.

Of course, recovering all original pixel colors from the quantized image is nearly impossible, thus our goal is to render a plausible version of what the original image might look like. The idea is to collect training data and train a ConvNet~\cite{LeCun-et-al-IEEE98,Ronneberger-etal-MICAI15,zhao2018identity,huang2018deep,natsume2018siclope} to map a quantized image to its original version. It is however difficult to obtain a good dequantization network for a wide range of GIF images.  To this end, we propose two novel techniques to improve the performance of the dequantization network. Firstly, we pose dequantization as an optimization problem, and we propose Compositional Color Dequantization Network (CCDNet), a novel network architecture for iterative color dequantization. Similar to the iterative Lucas-Kanade algorithm~\cite{Lucas-Kanade-IUW81}, this iterative procedure alleviates the problems associated with severe color quantization. Secondly, during training, we consider reconstruction loss and  generative adversarial loss~\cite{Goodfellow-etal-NIPS14,Radford-etal-arXiv15,Isola-etal-CVPR17} on both pixel colors and image gradients. This turns out to be far more effective than a loss function defined on the color values only.

Another contribution of the paper is the creation of two large datasets: GIF-Faces and GIF-Moments. Both datasets contain animated GIFs and their corresponding high-quality videos. GIF-Faces is face-centric whereas GIF-Moments is more generic and diverse.  Experiments on these two datasets demonstrate that our method can significantly enhance the visual quality of GIFs and reduce all types of artifacts. Comparatively, Our method outperforms its direct baselines as well as existing methods such as False Contour Detection \& Removal~\cite{huang2018fcdr} and Pix2Pix~\cite{Isola-etal-CVPR17}.

\section{Related Work}

\myheading{False Contour Detection and Removal.} Smooth areas of images and video frames should not contain color edges, but false contours are often visible in those areas after color bit-depth reduction or video codec encoding.  Several false contour detection and decontouring  methods~\cite{daly2004decontouring, ahn2005flatdetect, lee2006directcontrast, bhagavathy2009multiscale, yoo2009loop, jin2011composite,huang2018fcdr} have been proposed to address this problem. Among them, False Contour Detection and Removal (FCDR)~\cite{huang2018fcdr} is the most recent state-of-the-art approach. It first locates the precise positions of the false contours and then applies dedicated operations to suppress them. However, the color quantization artifacts in GIFs are far more severe, and GIF color dequantization requires more than removing minor false contours produced by bit-depth reduction.

\myheading{Video Interpolation.}  Classical video interpolation methods rely on cross-frame motion estimation and occlusion reasoning~\cite{herbst2009occlusion, barron1994performance, ilg2017flownet, baker2011database}. However, motion boundaries and severe occlusions are still challenging for existing optical flow estimation methods~\cite{geiger2012we, butler2012naturalistic}. Moreover, the flow computation, occlusion reasoning, and frame interpolation are separated steps that are not properly coupled. Drawing inspiration from the success of deep learning in high-level vision tasks~\cite{Krizhevsky-et-al-NIPS12, Simonyan-Zisserman-arXiv14,He-et-al-CVPR16}, many deep models have been proposed for single-frame interpolation~\cite{Liu-etal-ICCV17, long2016matching, Niklaus_CVPR_2017, Niklaus_ICCV_2017} and  multi-frame interpolation~\cite{superslomo}. SuperSlomo~\cite{superslomo} is a recently proposed state-of-the-art method for variable-length multi-frame interpolation approach. We adapt this method for GIF frame interpolation to enhance the temporal resolution of the input GIFs.

\section{GIF Generation and Artifacts} \label{sec: gif_generation}

The three main steps of creating a GIF from a video are: (1) frame sampling, (2) color quantization, and (3) color dithering. Frame sampling reduces the file size of the obtained GIF, but it also lowers the temporal resolution of the video content. In this section, we will provide more details about the color quantization and color dithering processes and the resulting visual artifacts as seen in Figure \ref{fig: gif_generation}.

\subsection{GIF Color Quantization}

The GIF color quantization process takes an input image $I \in R^{H\times W\times 3}$ and a color palette $\mathcal{C}\in R^{N \times 3}$ of $N$ distinct colors, and produces a color-quantized GIF image $G$. 
The quantization is computed for each pixel, thus $G$ has the same width and height as the input image.
$G_{i,j}$ at pixel $(i, j)$ is simply set to the color in the palette $\mathcal{C}$ closest to the input color $I_{i,j}$, i.e., $G_{i, j} = \myargmin{c ~\in~ \mathcal{C}}{~\norm{I_{i,j} - c}_2^2}$.
The color palette $\mathcal{C}$ could be optimized with a clustering algorithm to minimize the total quantization error $\norm{I-G}_2^2$. Different clustering algorithms are used in practice, but Median Cut~\cite{heckbert1982mediancut} is the most popular one due to its computational efficiency.

Most of the visual artifacts in GIFs are produced by the color quantization process with a tiny color palette ($N=256, 32, ...$). As illustrated in Figure \ref{fig: gif_generation}, the three most noticeable types of artifacts are (1) flat regions, (2) false contours, and (3) color shift. We notice a GIF image has a lot of connected components in which the color values are the same, which will be referred to as ``flat regions''. Flat regions are created because neighboring pixels with similar colors are quantized into the same color bin in the palette. False contours also emerge at the boundaries between flat regions with close color values. This is because the continuity of the color space has been broken and the color change cannot be gradual. We also notice the color shift between the input image and the GIF is larger for certain small regions such as the lips of the baby in Figure \ref{fig: gif_generation}. This is because the color palette does not spend budget on these small regions even though they have unique, distinct colors. 

\subsection{GIF Color Dithering}

Color quantization with a small color palette yields substantial quantization error and artifacts. Color dithering is a technique that can be used to hide quantization error and alleviate large-scale visual patterns such as false contours in GIFs. The most popular color dithering approach is Floyd-Steinberg dithering~\cite{floyd1976}. It diffuses the quantization error from every pixel to its neighboring pixels in a feedback process. The dithered GIF has the same exact small color palette. However, it appears to have more colors. The idea is to use a neighborhood of mixed colors to visually approximate a color that is not in the color palette.

Color dithering produces its own visual artifacts, which are noise-like dotted patterns. These dotted patterns are apparent  when one pays attention to local regions. This type of artifact is somewhat easier to remove, because the dithered GIFs preserve more color information with the help of neighboring pixels using the error-diffusion algorithm than the non-dithered GIFs. It is worth noting that, even with color dithering, GIFs still contain flat regions, false contours, and shifted colors.

\section{Our Approach}

Our method converts a sequence of GIF frames into a video that has a substantially higher visual quality. There are two main steps: color dequantization and frame interpolation. For color dequantization, we develop a new compositional ConvNet, inspired by the iterative optimization procedure of the Lucas-Kanade algorithm~\cite{Lucas-Kanade-IUW81}. This network is trained by combining reconstruction loss and generative adversarial loss on both the color values  and the image gradient vectors. After performing color dequantization, we apply an improved video frame interpolation method to increase the temporal resolution of the output video.

\subsection{Color Dequantization}

Let $G = f_\mathcal{C}(\hat{I})$ denote the color quantization function, where $G$ and $\hat{I}$ are the GIF image and the original input image respectively, and $\mathcal{C}$ is the color palette used for quantization. The goal of color dequantization is to recover the original image given the GIF image $G$ and the color palette~$\mC$, i.e., $I = f_\mathcal{C}^{-1}(G)$. However, the quantization function $f_\mathcal{C}$ is a many-to-one mapping, so color dequantization is an ill-posed problem.
Our proposed approach embeds the quantization function $f_\mathcal{C}$ itself into the compositional network, which provides valuable information to guide the learning and inference of the inverse function.

\subsubsection{Compositional Architecture}

Given the color quantized GIF image $G$ and the color palette $\mathcal{C}$, we seek image $I$ that is close to the ground truth image $\hat{I}$, and at the same time satisfies the color quantization constraint $f_\mathcal{C}(I)=G$. This can be formulated as an optimization problem, minimizing the reconstruction error between $I$ and $\hat{I}$ as well as between $f_\mathcal{C}(I)$ and $G=f_\mathcal{C}(\hat{I})$, i.e., 
\begin{equation}
\begin{split}
\min_{I} ~\norm{I-\hat{I}}_2^2 + \lambda \norm{f_\mathcal{C}(I)-G}_2^2.
\end{split}
\label{eq: two_loss}
\end{equation}
The first loss term is the reconstruction error between the recovered image and the ground truth image, which can be directly computed based on the output of the neural network and the target ground truth image. However, the second loss term $\norm{f_\mathcal{C}(I)-G}_2^2$ cannot be directly used as a proper loss for $I$ because the derivative of the quantization function with respect to the input image, $\frac{\partial f_\mathcal{C}(I)}{\partial I}$, is 0 almost everywhere. This is because the quantization process uses a tiny color palette. 

We propose to use Lucas-Kanade to iteratively optimize for the second loss term. In each iteration, we compute an update for the recovered image to further minimize the loss:
\begin{equation}
\begin{split}
\min_{\Delta I}~\norm{f_\mathcal{C}(I+\Delta I) - G}_2^2,
\end{split}
\label{equ: lk_loss}
\end{equation}
where $\Delta I$ is the update to the current estimation of the ground truth image $I$. The Lucas-Kanade algorithm assumes $f_\mathcal{C}(I+\Delta I)$ is a linear function of $\Delta I$ for small $\Delta I$, and it can be well approximated by the first order Taylor series expansion, i.e., $f_\mathcal{C}(I+\Delta I) \approx f_\mathcal{C}(I) + \frac{\partial f_\mathcal{C}(I)}{\partial I} \Delta I$. Thus, solving Equation (\ref{equ: lk_loss}) can be approximated by solving: 
\begin{equation}
\begin{split}
\min_{\Delta I} ~\norm{f_\mathcal{C}(I) + \frac{\partial f_\mathcal{C}(I)}{\partial I} \Delta I - G }_2^2.
\end{split}
\label{equ: linear_lk_loss}
\end{equation}
The above is a quadratic program with respect to $\Delta I$, and there is a closed-form solution for the optimal update: 
\begin{equation}
\begin{split}
\Delta I = \left(\frac{\partial f_\mathcal{C}(I)}{\partial I}\right)^{+} (G - f_\mathcal{C}(I)),\\
\end{split}
\label{equ: update_rule}
\end{equation}
where $^{+}$ denotes the pseudo-inverse operator. The Lucas-Kanade algorithm iterates between computing the above update value and updating the parameters: $I = I + \Delta I$. 

\begin{figure}[t]
\begin{center}
\includegraphics[width=\linewidth]{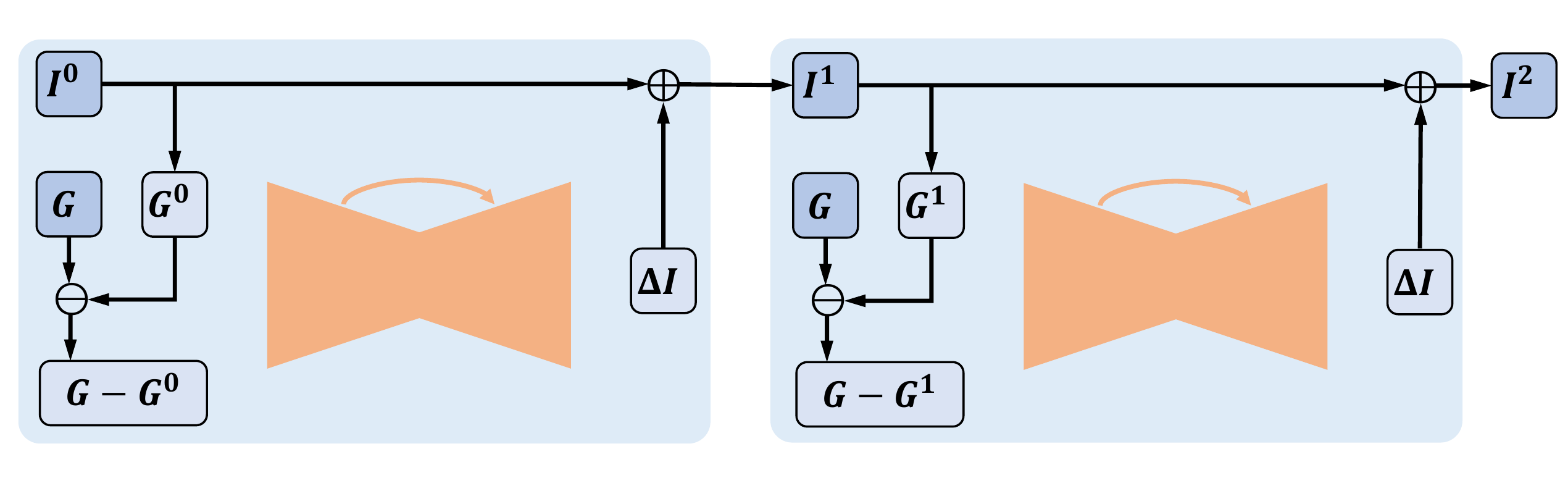}
\end{center}
\caption{{\bf Architecture of the proposed CCDNet.} Given the current image estimation $I^t$, we first compute its color quantified image $G^t$ using the same color palette of input GIF $G$. A UNet module then takes $(I^t, G, G^t, G-G^t)$ as input and outputs $\Delta I$, which will be added to the current image estimation. This process can be iteratively applied during training and test time.}
\label{fig: unfold2}
\end{figure}

Equation (\ref{equ: update_rule}) suggests the  update parameter  $\Delta I$ is a simple linear function of the difference between the two GIF images.  In practice, however, the true relationship between the quantization function and the input image is seldom linear. In this case, the linear approximation given by Taylor series expansion is not tight, and a simple linear model to compute the optimal update might not have enough capacity to fit the data. Instead, we propose to replace the linear function by a deep ConvNet. Specifically, we propose to use the U-Net architecture~\cite{Ronneberger-etal-MICAI15} to estimate the optimal update $\Delta I$. Equation \ref{equ: update_rule} becomes:
\begin{equation}
\begin{split}
\Delta I =  g(I, G, f_\mathcal{C}(I), G - f_\mathcal{C}(I)),
\end{split}
\label{equ: color_dequant1}
\end{equation}
where $g$ denotes the deep ConvNet that we need to learn. Following the iterative optimization scheme of the Lucas-Kanade algorithm, we alternate between computing the update direction and updating the de-quantized image: 
\begin{equation}
\begin{split}
&\Delta I = g(I^t, G, G^t, G - G^t), \\
&I^{t+1} \leftarrow ~ I^t + \Delta I 
\end{split}
\label{equ: color_dequant2}
\end{equation}
where $G^t=f_\mathcal{C}(I^t)$. This leads to the proposed Compositional Color Dequantization Network (CCDNet). The compositional architecture of the CCDNet is illustrated in Figure \ref{fig: unfold2}. Let $I^t$ be the current estimation of the ground truth image, we first apply $f_\mathcal{C}$ (the same color quantization function used to generate $G$) to $I^t$ to obtain $G^t$. Ideally $G^t$ should be identical to $G$. However if there is difference between the two quantified image, the difference $G - G^t$ will provide valuable information for estimating $\Delta I$ as shown in Equation \ref{equ: update_rule}. Therefore, we concatenate $(I^t,G,G^t,G-G^t)$ and apply network $g$ again to compute $\Delta I$, which is subsequently used to update the estimated image $I$. This process can be iteratively applied for multiple steps.

The CCDNet can be trained by unfolding the architecture multiple times, as illustrated in Figure \ref{fig: unfold2}. Suppose a CCDNet is unfolded by $k$ times, we refer to the corresponding model as CCDNet-$k$. Note that the same U-Net module is shared across all unfolding steps except for the first step. Reusing the same U-Net module dramatically reduces the number of model parameters compared to an alternative approach where the U-Net parameters at different stages are not shared. We allow the U-Net at the first unfolding step to have separate parameters from the rest because it expects different inputs ($I^t$ and $G^t$ are undefined for $t=0$).

For a color-dithered GIF, the exact quantization function $f_\mathcal{C}$ is unknown,  due to the missing information about the error diffusion step. Different GIF creation software programs use different error diffusion algorithms, and information about the algorithm is not stored in a GIF file. For color-dithered GIFs, we propose not to compute $G^t$ and $G-G^t$ in CCDNet. Fortunately, color-dithered and non-dithered GIFs have different local patterns, and they can easily recognized by a simple classifier. We propose to train two separate CCDNets for the color-dithered and non-dithered GIFs, and use a trained classifier to route an input GIF to the corresponding network. 

\begin{figure}[t]
\begin{center}
\includegraphics[width=\linewidth]{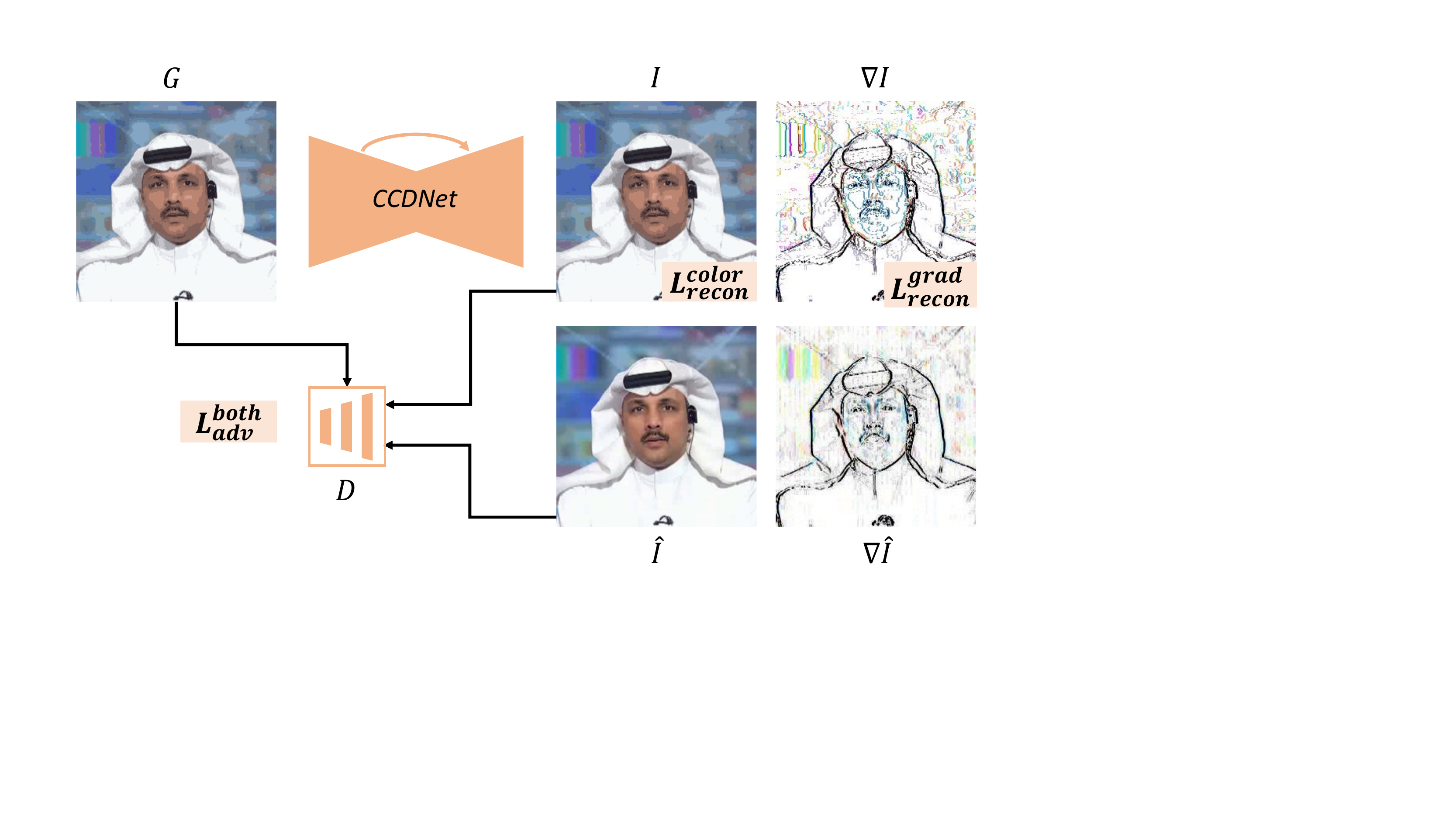}
\end{center}
\caption{{\bf Color Dequantization Loss.} The proposed loss in Equation \ref{equ: ccd_loss} measures the differences between the estimated image $I$ and the groundtruth image $\hat{I}$ based on both the color values and the gradient values. We can also train CCDNet using the conditional GAN framework to encourage even more realistic image outputs.}
\label{fig: ccd_loss}
\end{figure}

\subsubsection{Color Dequantization Loss}

Let $G_i$ be the input GIF image to the CCDNet and $I_i$ the corresponding output. We want to train a CCDNet so that $I_i$ is as close to the original image $\hat{I_i}$ as much as possible. We propose to use the loss function described in Equation (\ref{equ: ccd_loss}) to measure the discrepancy between these two images. 
\begin{align}
	%L_{total}(I_i, \hat{I_i}) &= 
	\mL_{recon}^{color}(I_i, \hat{I_i}) +  \mL_{recon}^{grad} (I_i, \hat{I_i})+  \lambda_{adv}  \mL_{adv}^{both}(I_i, \hat{I_i}).
    \label{equ: ccd_loss}
\end{align}
This loss function measures the differences between $I_i$ and $\hat{I_i}$ based on both the color values and the gradient values. To get sharper image estimation, we use $L_1$ norm to compute the reconstruction loss $\mL_{recon}^{color}$ and $\mL_{recon}^{grad}$:
\begin{align}
	\mL_{recon}^{color} = \norm{I_i - \hat{I_i}}_1, \textrm{and } 
\mL_{recon}^{grad} = \norm{\nabla I_i - \nabla \hat{I_i}}_1.
\end{align}
 We can also optimize the CCDNet using the conditional GAN framework, to encourage the outputs of the network to have the same distribution as the original ground truth images. This can be done by adding an adversarial loss function defined on both the color and gradient values: 
\begin{align}
	\mL_{adv}^{both} =  \log D(G_i, \hat{I_i}, \nabla \hat{I_i}) +  \log(1 - D(G_i, I_i, \nabla I_i)), \nonumber 
\end{align} 
where $D$ is the discriminator function for distinguishing between the set of ground truth images $\{\hat{I}_i\}$ and the set of estimated images $\{I_i\}$. $\lambda_{adv}$ is set to $0.01$ or $0$, depending on whether the adversarial loss is enabled. Experiments show it is critical to include the losses computed on the image gradient values. Compared to the original images, GIF images have drastically different gradient signatures (due to flat regions, false contours, dotted patterns), so it is much more effective to use additional losses on the image gradients.

%\begin{equation}
%\begin{split}
%L_{total} &= L_{recon}^{color} +  L_{recon}^{grad} +  \lambda_{adv}  L_{adv}^{both} \\
%L_{recon}^{color} &= \frac{1}{N}\sum_{i=1}^N \norm{I_i - \hat{I_i}}_1 \\
%L_{recon}^{grad} &= \frac{1}{N}\sum_{i=1}^N \norm{\nabla I_i - \nabla \hat{I_i}}_1 \\
%L_{adv}^{both} &=  \frac{1}{N}\sum_{i=1}^N \log D(G_i, \hat{I_i}, \nabla \hat{I_i})\\
%&+ \frac{1}{N}\sum_{i=1}^N \log(1 - D(G_i, I_i, \nabla I_i))
%\end{split}
%\label{equ: ccd_loss}
%\end{equation}

\subsection{Temporal Interpolation}

\begin{figure}[t]
\begin{center}
\includegraphics[width=\linewidth]{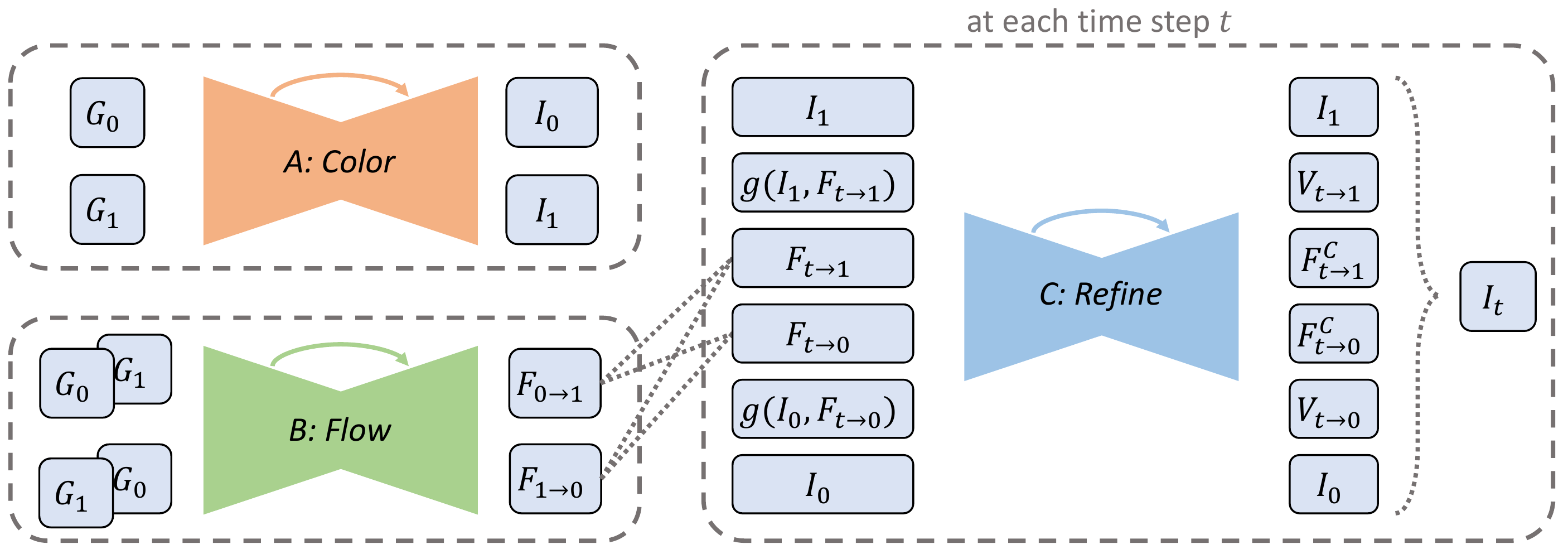}
\end{center}
  \caption{{\bf Overview of the GIF2Video pipeline.} Network A performs color dequantization on two consecutive input GIF frames $G_0$ and $G_1$; Network B estimates the bidirectional flow maps between them; Network C receives the outputs from A and B, and produces the interpolated frames $I_t$'s for $t\in (0, 1)$. We use the proposed CCDNet2 as Network A and the modified SuperSlomo as Network B and C.}
\label{fig: gif2video}
\end{figure}

We adapt the recently proposed SuperSlomo~\cite{superslomo} network to reverse the process of frame sampling and increase the temporal resolution of GIFs.  SuperSlomo is designed for variable-length multiple-frame interpolation. Given two consecutive frames at time steps $t=0$ and $t=1$, SuperSlomo in one step can interpolate frames anywhere between $t=0$ and $t=1$. This is more efficient than methods where only the middle frame $t=0.5$ is produced. More details about SuperSlomo can be found in~\cite{superslomo}. 

We implement SuperSlomo and adapt it to our task. Figure \ref{fig: gif2video} depicts the entire GIF2Video pipeline, with the adaptation is shown in (B) and (C). This algorithm has three major components. Network A performs color dequantization and outputs the estimated ground truth images $I_0$ and $I_1$. Network B estimates the bidirectional flow maps $F_{0\rightarrow 1}$ and $F_{1\rightarrow 0}$ between the two input frames. Network C receives the outputs of network A and B, and it produced interpolated frames $I_t$'s for $t \in (0, 1)$. We use the proposed CCDNet as network A, while network B and C are both U-Net modules from ~\cite{superslomo}. Note that network B estimates the optical flow directly from the input GIF images, instead of using the outputs of network A. This allows networks A and B to run in parallel. Experiments show this parallel model performs similarly to the alternative sequential model.

% \subsection{Implementation Details}
% \mtodo{describe the implementation details.}

\section{Datasets}

\begin{figure}[t]
\begin{center}
\includegraphics[width=0.85\linewidth]{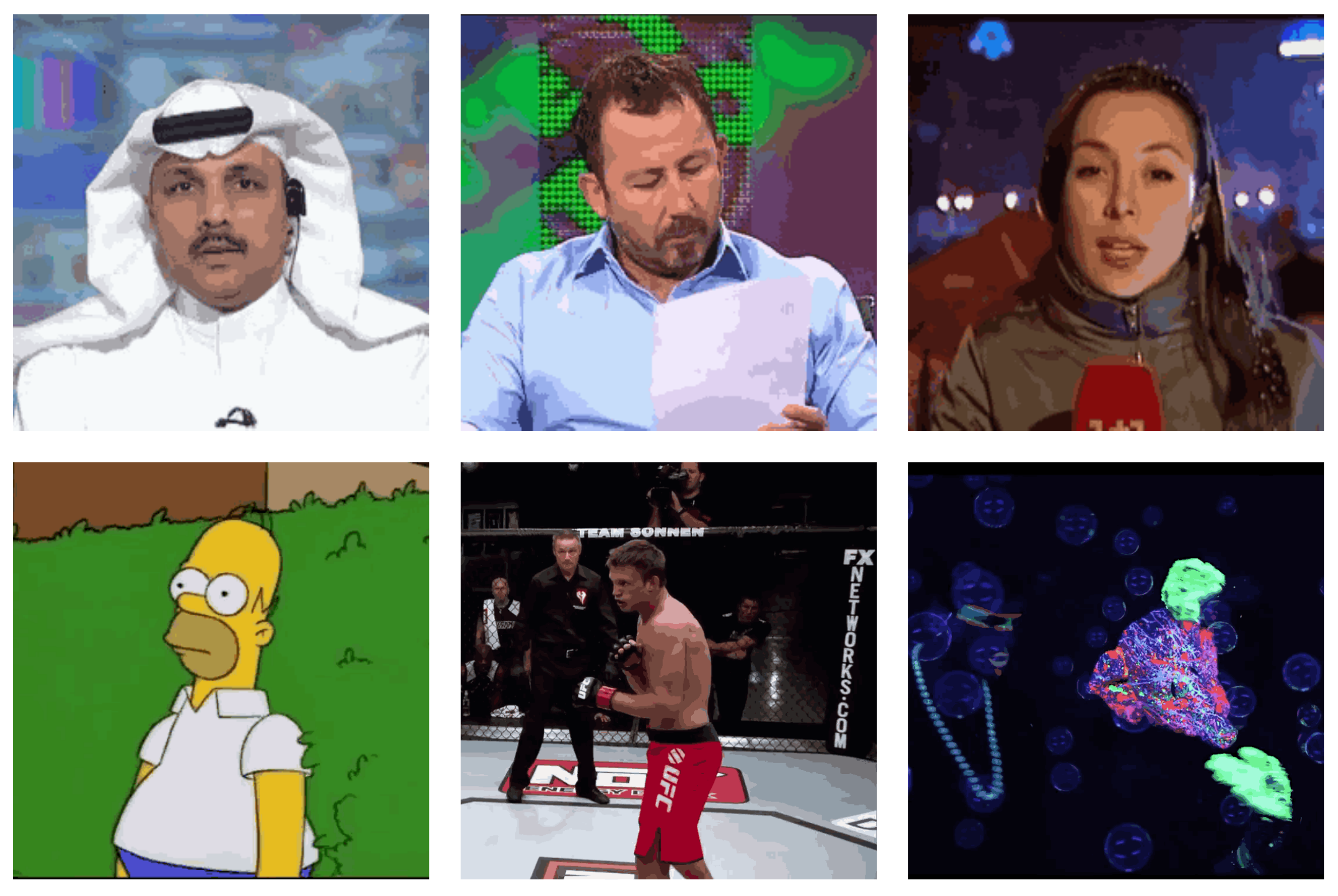}
\end{center}
   \caption{{\bf Example GIF frames from GIF-Faces and GIF-Moments.} First row: GIF-Faces (face-centric); Second row: GIF-Moments (generic GIFs shared by Internet users).}
\label{fig: gif_faces_moments}
\end{figure}

With methods presented in Section \ref{sec: gif_generation}, we can convert any video frame to a GIF image. This allows us to train the CCDNet with a large amount of training image pairs. As a byproduct, we introduce two GIF-Video datasets: GIF-Faces and GIF-Moments. The former is designed to be face-centric, while the latter is more generic and built on real GIFs shared by Internet users. Figure \ref{fig: gif_faces_moments} shows some GIF frames (non-dithered) of the two datasets. Images in the first row are from the GIF-Faces dataset, and they also cover parts of the upper-body with diverse background scenes. The second row shows images from the GIF-Moments dataset. They contain diverse content, covering a wide range of scenes including sports, movie, and animation. Details about these two datasets are provided below.

\subsection{GIF-Faces}

A large portion of online GIFs are face-centric, and they contain noticeable artifacts on face regions such as cheeks and lips. Given the popularity of face-centric GIFs, there are strong motivations for optimizing the network on faces. The GIF-Faces dataset was designed for such a purpose. 
%
%GIF-Faces is the first GIF-Video dataset we created for training the CCDNet. We also use this dataset for comprehensive evaluation and analysis of different algorithms. It is designed to be structured and face-centric. The reason is that we observe a large percentage of online GIFs are face-centric and the visual artifacts on faces skin and lips are very significant and representative.

We first extracted a large number of face-centric video clips from the FaceForensics dataset~\cite{rossler2018faceforensics}. Most of the faces in FaceForensics have near-frontal poses and neutral expression changes across frames. Given a video from FaceForensics, we first detected human faces on every frame, then computed a minimal square region which covers all the detected faces. We further expanded this square region by a factor of 1.5 to increase the coverage of the scene. 

A total of 987 high-quality face-centric video clips were extracted from the FaceForensics dataset. The frames of these videos were resized to have \htimesw{256p}{256p} resolution and the temporal resolution was unaltered. We used 883 videos for training and 104 for evaluation. There are around 500 frames on average in each video. The corresponding GIF frames (both dithered and non-dithered) were computed from these face-centric videos with the color palette size set to $32$. We use $32$ as the default color palette size, to make the color dequantization task as challenging as possible, yet not unreasonable. To deal with GIFs of different palette sizes, we can simply read their palette sizes and route them to appropriate models trained on similar levels of color quantization.

\subsection{GIF-Moments}

We also curated GIF-Moments, a dataset of generic GIFs shared by Internet users. Specifically, \citet{gygli2016video2gif} crawled popular GIF-sharing websites and collected 100K GIFs and their corresponding original videos. For each GIF clip, their dataset provides the corresponding YouTube video ID and the start frame and end frame. These video moments are generic and diverse, covering a wide range of video categories and topics such as sports, movie, and animation. We first downloaded all the candidate videos from YouTube in their highest resolution, and temporally cropped the videos using the annotated start and end frames. We only kept videos of high visual quality with sufficient spatial and temporal resolution: the width and height must be at least $360p$, the temporal resolution is no less than 20fps, and the total number of frames has to be more than 40.

In the end, we had a collection of 71,575 video clips, with a total of 12 million frames. We use 85\%, 5\%, and 10\% of the videos for training, validation, and evaluation respectively. Similar to GIF-Faces, we computed the corresponding GIF frames (both dithered and non-dithered) with the color palette size set to $32$.

\section{Experiments}

In our experiments, PSNR (Peak Signal to Noise Ratio) and SSIM (Structural Similarity Index) are used as evaluation metrics. PSNR is defined via the root mean squared error (RMSE) between the estimated images and the ground truth images. More specifically, $PSNR = 20 \log_{10}\frac{MAX}{RMSE}$. Roughly speaking, 1dB, 2dB, and 3dB improvement on the PSNR are equivalent to 10\%, 20\%, and 30\% RMSE reduction in the image color space respectively. SSIM is a perceptual metric that quantifies image quality. We first compute PSNR and SSIM for each frame and average them within each video, and finally average them across all videos in the test set.

\subsection{GIF Color Dequantization}

\subsubsection{Dichotomy of GIF Color Dithering Mode}

\begin{figure}[t]
\begin{center}
\includegraphics[width=0.85\linewidth]{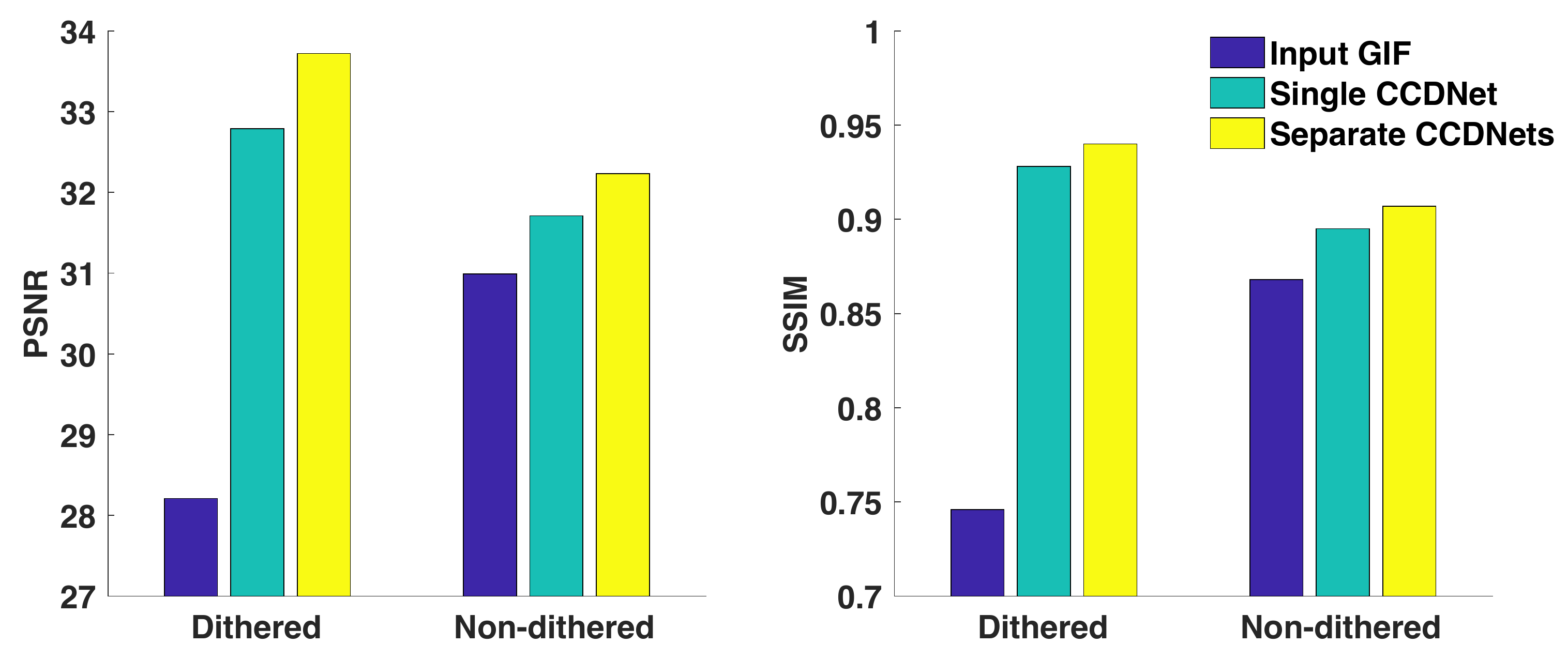}
\end{center}
\vskip -0.1in
   \caption{{\bf Benefits of having separate dequantization networks for color-dithered and non-dithered GIFs}. Cyan: using a single CCDNet for both types of images. Yellow: using separate dedicated networks. Here we use the GIF-Faces dataset to train CCDNet1 with $\lambda_{adv}=0.01$. The performance is measured by PSNR and SSIM (higher is better).}   
\label{fig: single-or-separate}
\end{figure}

The color dithering step is optional in the GIF generation process. It is up to the creation tools or the end users to decide whether or not to perform color dithering after color quantization. We observe that color-dithered GIFs are more popular than the non-dithered ones on the Internet. Color-dithering is preferred because it can reduce the noticeability of large visual artifacts such as false contours in GIFs. However, non-dithered GIFs are also widely shared. They also exhibit more artifacts and are more challenging for the task of GIF color dequantization.

Should we learn a single CCDNet or two separate CCDNets for the color-dithered and the non-dithered GIFs? The latter approach is better, as suggested by Figure \ref{fig: single-or-separate}. This figure shows the results of an experiment on the GIF-Faces dataset, where we train a CCDNet1 model with $\lambda_{adv} = 0.01$. From Figure \ref{fig: single-or-separate}, we observe that learning a single CCDNet with both dithered and non-dithered GIFs used as training data reduces the GIF color dequantization performance, measured by PSNR and SSIM.
We also observe that it is easier to restore color values for the dithered GIFs. The reason is that dithered GIFs preserve more color information than non-dithered ones with the help of neighboring pixels using error-diffusion algorithms. 

%So it is more challenging to perform color dequantization for the non-dithered GIFs.

The benefits of having separate CCDNets for color-dithered and non-dithered GIFs is clear. However, at test time, the color-dithering mode of a GIF image is not stored in the file. Fortunately, the color-dithered and non-dithered GIFs exhibit very different local patterns especially on the gradient space. We therefore can train a classifier to infer whether an input GIF is dithered or not. We trained a simple classifier with only five Conv layers on the GIF-Faces training set. It achieves 100\% and 98.6\% accuracy on the GIF-Faces and GIF-Moments test sets respectively. 
The model is a CNN with layers: $C(9, 64)\rightarrow NL\rightarrow C(64, 128)\rightarrow NL \rightarrow C(128, 256) \rightarrow NL \rightarrow C(512, 1) \rightarrow GAP$. $C(m,n)$ represents a Conv layer with $m$ input channels and $n$ output channels. $NL$ stands for Non-Linearity, which is one BatchNorm followed by one LeakyReLU (negative slope: 0.2). $GAP$ is short for Global Average Pooling. The input is a GIF frame with its gradient maps.

\subsubsection{Ablation Study on Networks \& Losses}

\begin{table}[t]
\begin{center}
\begin{tabular}{r|lcc}
\toprule
\multirow{2}{*}{\#} & \multirow{2}{*}{Method}& \multicolumn{2}{c}{GIF-Faces} \\
% \cmidrule(lr){3-4}
 &  & Non-dithered & Dithered \\
\midrule
\midrule
1& DRRN-1 & 30.52/0.874 & - \\
2& GLCIC-1 & 30.71/0.883 & - \\
3& UNet-1 & \textbf{32.23/0.907} & \textbf{33.72/0.940} \\
\midrule
4& UNet-1 (no grad loss) &  31.20/0.884 & 32.68/0.927 \\
5& UNet-1 (no adv loss) & \textbf{32.83/0.918} & \textbf{33.90/0.944} \\
\midrule
6& UNet-2  & 32.65/0.911 & 34.31/0.943 \\
7& UNet-3  & \textbf{32.85/0.917} & \textbf{34.43/0.945} \\
\midrule
8& UNet-2 (no adv loss) & \textbf{34.05/0.928} & \textbf{35.63/0.956} \\
9& UNet-3 (no adv loss) & 33.75/0.927 & 34.59/0.950 \\
\midrule
\midrule
10& Pix2Pix~\cite{Isola-etal-CVPR17}  &  31.41/0.895 & 32.80/0.925 \\
11& FCDR~\cite{huang2018fcdr}  & 31.51/0.878 & - \\
12& Gaussian ($\sigma=0.5$)  & 31.36/0.876 & 31.20/0.873 \\
13& Gaussian ($\sigma=1.0$)  & 29.26/0.797 & 29.75/0.815 \\
\midrule
14& GIFs (palette: 32) & 30.99/0.868 & 28.21/0.746 \\
\bottomrule
\end{tabular}
\end{center}
    \caption{{\bf Quantitative Results of GIF Color Dequantization on GIF-Faces.} Row 1-9 are the results of CCDNet with different settings. UNet-$k$ stands for CCDNet-$k$ with UNet as backbone. Row 10-13 are the results of several existing methods. 
    The performance is measured by PSNR and SSIM (higher is better).
    }
\label{tab: all_in_one}
\end{table}

We perform extensive ablation study and quantitative evaluation on the GIF-Faces dataset. From these experiments, we draw several conclusions below.

\myheading{U-Net is an effective building block for CCDNet.} Function $g$ in Equation (\ref{equ: color_dequant2}) denotes a deep neural network for computing the iterative update. There are many candidate network architectures that can be used for $g$. We experiment with three models that have been successfully used for other tasks that are similar to color dequantization:  U-Net~\cite{Ronneberger-etal-MICAI15}, DRRN~\cite{Tai-DRRN-2017}, and GLCIC~\cite{iizuka2017glcic}. The U-Net architecture allows multi-level information to shortcut across the network and is widely used for the task of image segmentation and image-to-image translation~\cite{Isola-etal-CVPR17}. DRRN (Deep Recursive Residual Network) is a state-of-the-art network for single image super-resolution. It can substantially reduce the amount of model parameters by applying residual learning in both global and local manners. GLCIC (Globally and Locally Consistent Image Completion) is proposed for the task of image and video inpainting~\cite{wang2018videoinp}. The mid-layers of GLCIC are dilated Conv layers~\cite{yu2015multi}, allowing to compute each output pixel with a much larger input area without increasing the amount of model parameters. 

\begin{figure}[t]
\begin{center}
\includegraphics[width=0.85\linewidth]{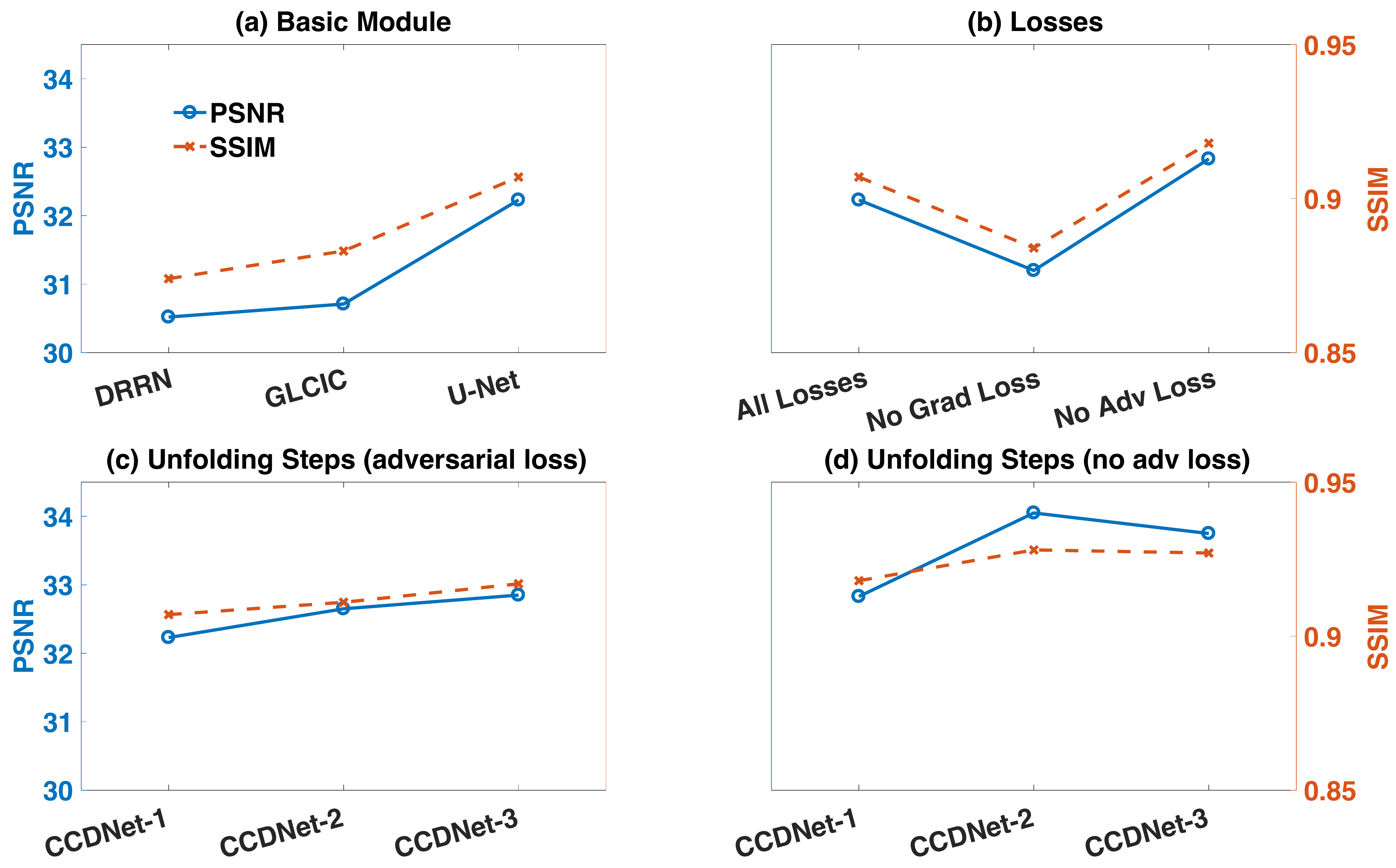}
\end{center}
   \caption{{\bf Ablation study of CCDNet on GIF-Faces dataset.} (a) U-Net is a more effective building block for CCDNet than others. (b) It is critical to include the loss defined on the gradient values, and using adversarial loss yields more realistic images. (c, d) It is beneficial to unfold CCDNet by multiple steps.}
\label{fig: ablation_face}
\end{figure}

The results are shown in Table \ref{tab: all_in_one} (Row 1-3) and  Figure \ref{fig: ablation_face} (a). As can be observed, using U-Net as the basic module for CCDNet is significantly better than using DRRN or GLCIC. We believe that DRRN's capability of recovering colors from severely quantized GIFs is limited by its small parameter size. And GLCIC is generally bad at predicting image within regions of high-frequency textures (e.g., stripes on clothes, text on background).

\setlength{\tabcolsep}{3pt}
\begin{table}[t]
\begin{center}
\begin{tabular}{lcccc}
\toprule
{\small  Sample}  & \multicolumn{2}{c}{GIF-Faces} & \multicolumn{2}{c}{GIF-Moments} \\
  \cmidrule(lr){2-3} \cmidrule(lr){4-5}
Rate & GIF & GIF2Video & GIF & GIF2Video  \\
\midrule
1/1 & 30.99/0.868 & \textbf{34.05/0.928} & 33.71/0.902 & \textbf{36.10/0.948}  \\
1/2 & 30.02/0.857 & \textbf{33.27/0.921} & 29.05/0.859 & \textbf{31.92/0.918} \\
1/4 & 29.01/0.842 & \textbf{32.08/0.908} & 26.16/0.812 & \textbf{28.38/0.865} \\
1/8 & 27.41/0.815 & \textbf{30.20/0.884} & 23.29/0.751 & \textbf{24.95/0.800}  \\
\bottomrule
\end{tabular}
\end{center}
    \caption{{\bf Results of temporal GIF frame interpolation.} The visual quality of created GIFs quickly deteriorates as the temporal downsampling factor increases from 1 to 8. The proposed GIF2Video improves the PSNR of recovered videos by 3dB on GIF-Faces dataset, that is equivalent to 30\% root-mean-square-error reduction on the pixel color values.}
\label{tab: temporal_interp}
\end{table}

%\begin{table}[t]
%\begin{center}
%\begin{tabular}{c|cc|cc}
%\toprule
%Temporal  & \multicolumn{2}{c|}{GIF-Faces} & \multicolumn{2}{c}{GIF-Moments} \\
%Downsampling & GIF & GIF2Video & GIF & GIF2Video  \\
%\midrule
%1 & 30.99/0.868 & \textbf{34.05/0.928} & 33.71/0.902 & \textbf{36.10/0.948}  \\
%2 & 30.02/0.857 & \textbf{33.27/0.921} & 29.05/0.859 & \textbf{31.92/0.918} \\
%4 & 29.01/0.842 & \textbf{32.08/0.908} & 26.16/0.812 & \textbf{28.38/0.865} \\
%8 & 27.41/0.815 & \textbf{30.20/0.884} & 23.29/0.751 & \textbf{24.95/0.800}  \\
%\bottomrule
%\end{tabular}
%\end{center}
%    \caption{{\bf Results of temporal GIF frame interpolation.} The visual quality of created GIFs quickly deteriorates as the temporal downsampling factor increases from 1 to 8. The proposed GIF2Video improves the PSNR of recovered videos by 3dB on GIF-Faces dataset, that is equivalent to 30\% root-mean-square-error reduction on the pixel color values.}
%\label{tab: temporal_interp}
%\end{table}

%  

\begin{figure}[t]
\begin{center}
\includegraphics[width=\linewidth]{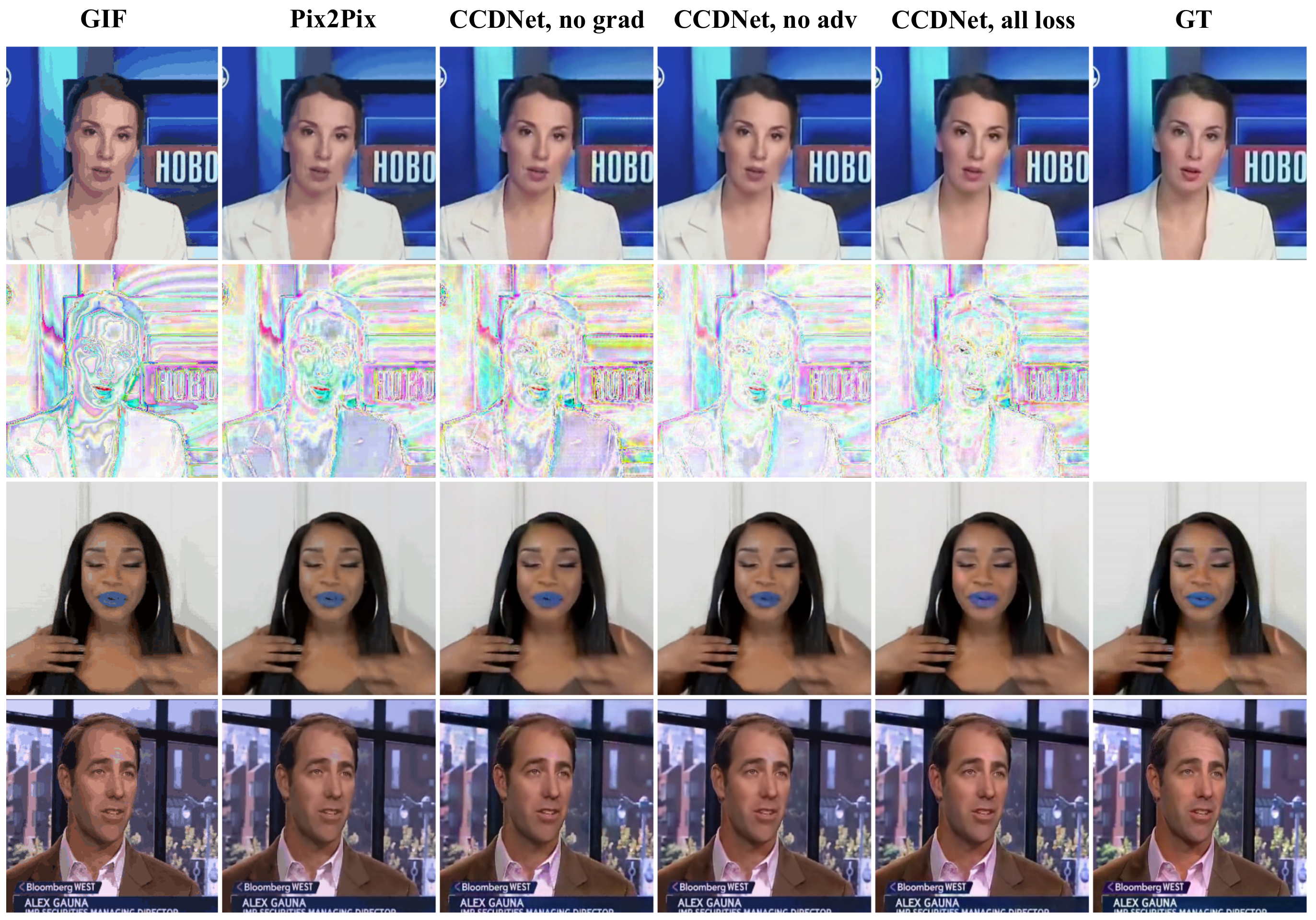}
\end{center}
\caption{{\bf Qualitative Results of GIF Color Dequantization on GIF-Faces.} Pix2Pix and CCDNet trained without image gradient-based losses cannot remove quantization artifacts such as flat regions and false contours very well. Training CCDNet with adversarial loss yields more realistic and colorful images (see the color of the skin and the lips). Best viewed on a digital device. 
}
\label{fig: qualitative}
\end{figure}

\myheading{It is critical to include the loss defined on the gradient values.} Row 3 of Table \ref{tab: all_in_one} shows the color dequantization performance achieved by training CCDNet1 with all loss terms, whereas Row 4 shows the performance when the loss on the gradient values is excluded. More specifically, to disable the loss of the gradient values, we discard the gradient reconstruction loss $\mL_{recon}^{grad}$ and stop using $\nabla I$ as input channels to the adversarial discriminator $D$. Comparing Row 3 and 4 (also see Figure\ref{fig: ablation_face}(b)), we observe a significant quantitative performance drop after disabling the image gradient-based losses. But more importantly, we find that without the gradient-based losses, the network fails to reduce the artifacts such as flat regions and false contours. The reason can be seen from Figure \ref{fig: ccd_loss}: the difference between $I$ and $\hat{I}$ is more apparent in the gradient space than in the original color space. Because the artifacts exhibited in GIFs have drastically different gradient signatures from the ground truth images. For example, the flat regions in GIFs have zero image gradient while the dotted patterns in GIFs exhibit noise-like gradient patterns.

\myheading{Using adversarial loss yields more realistic images.} Comparing Row 3 and 5 of Table~\ref{tab: all_in_one} and also considering Figure\ref{fig: ablation_face}(b), we observe that after removing the adversarial loss $\mL_{adv}^{both}$, the quantitative performance measured by PSNR/SSIM actually improves. This is not surprising, as it is aligned with many previous studies using adversarial learning. The adversarial loss is designed to make the output images more realistic, which is sometimes not perfectly aligned with the goal of improving quantitative measures such as PSNR or SSIM. Looking at qualitative results, we find that the adversarial loss is indeed helpful to make the output images more realistic. We also performed a small scale user study involving five subjects. We displayed a pair of images produced by the two CCDNet2 (with and without $\mL_{adv}$) in randomized order. The subjects chose which is more realistic to them. The three choices and the percentage being chosen are as follows. \textit{Use $\mL_{adv}$}: 53\%; \textit{No $\mL_{adv}$}: 12\%; \textit{Not Sure}: 35\%. 

\myheading{It is beneficial to unfold CCDNet by multiple steps and embed the quantization process into the CCDNet.} As illustrated in Figure \ref{fig: unfold2}, the proposed CCDNet is a compositional architecture that can be unfolded by multiple steps. Empirically, it is beneficial to do so, as can be seen from Figure \ref{fig: ablation_face}(c) and (d). We observe that with more unfolding steps, the CCDNet can estimate the ground truth image more accurately especially around the object boundaries. We also investigate if it is effective to embed the GIF color quantization process into the CCDNet. In Equation \ref{equ: update_rule}, \ref{equ: color_dequant1} and \ref{equ: color_dequant2}, we derive that the difference image between the input GIF and the corresponding GIF of the current estimation, i.e., $G - G^t = G - f_{\mathcal{C}}(I^t)$, provides valuable information and guidance on how to update the current estimation. If we remove $G^t$ and $G-G^t$ from the input channels to the U-Net basic module, the color dequantization performance of CCDNet2 will decrease significantly. For CCDNet2 trained without $\mL_{adv}$, PSNR/SSIM drops from 34.05/0.956 to 33.40/0.923. For CCDNet2 trained with $\mL_{adv}$, PSNR/SSIM drops from 32.65/0.911 to 32.48/0.904.

\subsubsection{Comparison to Other Methods}

Table \ref{tab: all_in_one} Row 10-13 report the color dequantization performance of several other methods on GIF-Faces dataset. We first consider applying Gaussian Smoothing with different kernel sizes (Row 12, 13). As expected, the color dequantization performance of this naive approach is really poor. We then implement FCDR (False Contour Detection \& Removal~\cite{huang2018fcdr}), a recently proposed state-of-the-art method for image bit-depth superresolution. It can alleviate mild color quantization artifacts introduced by image bit-depth reduction. However, the color quantization used in GIF generation is far more aggressive than that in image bit-depth reduction. FCDR cannot handle severe GIF artifacts, as listed in Row 11. We also tested Pix2Pix~\cite{Isola-etal-CVPR17}, an adversarial network designed for image-to-image translation tasks. It performs similarly to our CCDNet1 trained without image gradient-based losses.

\subsection{Temporal GIF Frame Interpolation}

Table \ref{tab: temporal_interp} shows the performance of the proposed GIF2Video algorithm on non-dithered GIF-Faces and GIF-Moments datasets. The performance is measured by PSNR/SSIM (higher is better). For this experiment, we use CCDNet2 trained without the adversarial loss for color dequantization. As can be observed, the visual quality of created GIFs quickly deteriorates as the temporal downsampling factor increases from 1 to 8. For a large downsampling factor, the visual quality of GIF-Moments is worse than that of GIF-Faces. This is because the GIF-Moments dataset contains more dynamic content and larger motions. With the proposed GIF2Video algorithm, we are able to improve the PSNR of recovered videos by 3dB on GIF-Faces dataset, that is equivalent to 30\% root-mean-square-error reduction in the image color space.

\section{Conclusions}
This paper presents GIF2Video, the first learning-based method  for enhancing the visual quality of GIFs in the wild. The main tasks of GIF2Video are color dequantization and frame interpolation. For the first task, we propose a novel compositional network architecture CCDNet and a comprehensive loss for training it. 
For the second task, we adapt SuperSlomo for variable-length multi-frame interpolation to enhance the temporal resolution of input GIFs. 
Experiments show our method can dramatically enhance the visual quality of input GIFs and significantly reduce quantization artifacts.
We hope our method could inspire more solutions to the task of reconstructing video from GIF, such as based on the idea of viewing image sequences as a 3D volume~\cite{wang2014video,wang2017video}, or applying recurrent neural networks to enhance the inter-frame consistency~\cite{ren2016look}.

{\small
\bibliographystyle{abbrvnat}
\bibliography{shortstrings,pubs,mybib}

\begin{thebibliography}{44}
\providecommand{\natexlab}[1]{#1}
\providecommand{\url}[1]{\texttt{#1}}
\expandafter\ifx\csname urlstyle\endcsname\relax
  \providecommand{\doi}[1]{doi: #1}\else
  \providecommand{\doi}{doi: \begingroup \urlstyle{rm}\Url}\fi

\bibitem[gif()]{giflink}
Graphics interchange format, version 89a.
\newblock \url{https://www.w3.org/Graphics/GIF/spec-gif89a.txt}.

\bibitem[Ahn and Kim(2005)]{ahn2005flatdetect}
W.~Ahn and J.-S. Kim.
\newblock Flat-region detection and false contour removal in the digital tv
  display.
\newblock In \emph{Multimedia and Expo, 2005. ICME 2005. IEEE International
  Conference on}, pages 1338--1341. IEEE, 2005.

\bibitem[Baker et~al.(2011)Baker, Scharstein, Lewis, Roth, Black, and
  Szeliski]{baker2011database}
S.~Baker, D.~Scharstein, J.~Lewis, S.~Roth, M.~J. Black, and R.~Szeliski.
\newblock A database and evaluation methodology for optical flow.
\newblock \emph{IJCV}, 92\penalty0 (1):\penalty0 1--31, 2011.

\bibitem[Barron et~al.(1994)Barron, Fleet, and
  Beauchemin]{barron1994performance}
J.~L. Barron, D.~J. Fleet, and S.~S. Beauchemin.
\newblock Performance of optical flow techniques.
\newblock \emph{IJCV}, 12\penalty0 (1):\penalty0 43--77, 1994.

\bibitem[Bhagavathy et~al.(2009)Bhagavathy, Llach, and
  Zhai]{bhagavathy2009multiscale}
S.~Bhagavathy, J.~Llach, and J.~Zhai.
\newblock Multiscale probabilistic dithering for suppressing contour artifacts
  in digital images.
\newblock volume~18, pages 1936--1945. IEEE, 2009.

\bibitem[Butler et~al.(2012)Butler, Wulff, Stanley, and
  Black]{butler2012naturalistic}
D.~J. Butler, J.~Wulff, G.~B. Stanley, and M.~J. Black.
\newblock A naturalistic open source movie for optical flow evaluation.
\newblock In \emph{Proc. ECCV}, pages 611--625. Springer, 2012.

\bibitem[Daly and Feng(2004)]{daly2004decontouring}
S.~J. Daly and X.~Feng.
\newblock Decontouring: Prevention and removal of false contour artifacts.
\newblock In \emph{Human Vision and Electronic Imaging IX}, volume 5292, pages
  130--150. International Society for Optics and Photonics, 2004.

\bibitem[Floyd and Steinberg(1976)]{floyd1976}
R.~Floyd and L.~Steinberg.
\newblock Adaptive algorithm for spatial greyscale.
\newblock In \emph{Proceedings of the Society of Information Display}, 1976.

\bibitem[Geiger et~al.(2012)Geiger, Lenz, and Urtasun]{geiger2012we}
A.~Geiger, P.~Lenz, and R.~Urtasun.
\newblock Are we ready for autonomous driving? the kitti vision benchmark
  suite.
\newblock In \emph{Proc. CVPR}, pages 3354--3361. IEEE, 2012.

\bibitem[Goodfellow et~al.(2014)Goodfellow, Pouget-Abadie, Mirza, Xu,
  Warde-Farley, Ozair, Courville, and Bengio]{Goodfellow-etal-NIPS14}
I.~Goodfellow, J.~Pouget-Abadie, M.~Mirza, B.~Xu, D.~Warde-Farley, S.~Ozair,
  A.~Courville, and Y.~Bengio.
\newblock Generative adversarial nets.
\newblock In Z.~Ghahramani, M.~Welling, C.~Cortes, N.~D. Lawrence, and K.~Q.
  Weinberger, editors, \emph{NIPS}. 2014.

\bibitem[Gygli et~al.(2016)Gygli, Song, and Cao]{gygli2016video2gif}
M.~Gygli, Y.~Song, and L.~Cao.
\newblock Video2gif: Automatic generation of animated gifs from video.
\newblock In \emph{Proc. CVPR}, 2016.

\bibitem[He et~al.(2016)He, Zhang, Ren, and Sun]{He-et-al-CVPR16}
K.~He, X.~Zhang, S.~Ren, and J.~Sun.
\newblock Deep residual learning for image recognition.
\newblock In \emph{Proc. CVPR}, 2016.

\bibitem[Heckbert(1982)]{heckbert1982mediancut}
P.~Heckbert.
\newblock \emph{Color image quantization for frame buffer display}, volume~16.
\newblock ACM, 1982.

\bibitem[Herbst et~al.(2009)Herbst, Seitz, and Baker]{herbst2009occlusion}
E.~Herbst, S.~Seitz, and S.~Baker.
\newblock Occlusion reasoning for temporal interpolation using optical flow.
\newblock \emph{Department of Computer Science and Engineering, University of
  Washington, Tech. Rep. UW-CSE-09-08-01}, 2009.

\bibitem[Huang et~al.(2018{\natexlab{a}})Huang, Kim, Tsai, Jeong, Choi, and
  Kuo]{huang2018fcdr}
Q.~Huang, H.~Y. Kim, W.-J. Tsai, S.~Y. Jeong, J.~S. Choi, and C.-C.~J. Kuo.
\newblock Understanding and removal of false contour in hevc compressed images.
\newblock \emph{IEEE Transactions on Circuits and Systems for Video
  Technology}, 28\penalty0 (2):\penalty0 378--391, 2018{\natexlab{a}}.

\bibitem[Huang et~al.(2018{\natexlab{b}})Huang, Li, Chen, Zhao, Xing, LeGendre,
  Luo, Ma, and Li]{huang2018deep}
Z.~Huang, T.~Li, W.~Chen, Y.~Zhao, J.~Xing, C.~LeGendre, L.~Luo, C.~Ma, and
  H.~Li.
\newblock Deep volumetric video from very sparse multi-view performance
  capture.
\newblock In \emph{Proceedings of the European Conference on Computer Vision
  (ECCV)}, pages 336--354, 2018{\natexlab{b}}.

\bibitem[Iizuka et~al.(2017)Iizuka, Simo-Serra, and Ishikawa]{iizuka2017glcic}
S.~Iizuka, E.~Simo-Serra, and H.~Ishikawa.
\newblock Globally and locally consistent image completion.
\newblock \emph{ACM Transactions on Graphics (TOG)}, 36\penalty0 (4):\penalty0
  107, 2017.

\bibitem[Ilg et~al.(2017)Ilg, Mayer, Saikia, Keuper, Dosovitskiy, and
  Brox]{ilg2017flownet}
E.~Ilg, N.~Mayer, T.~Saikia, M.~Keuper, A.~Dosovitskiy, and T.~Brox.
\newblock Flownet 2.0: Evolution of optical flow estimation with deep networks.
\newblock In \emph{Proc. CVPR}, volume~2, page~6, 2017.

\bibitem[Isola et~al.(2017)Isola, Zhu, Zhou, and Efros]{Isola-etal-CVPR17}
P.~Isola, J.-Y. Zhu, T.~Zhou, and A.~A. Efros.
\newblock Image-to-image translation with conditional adversarial networks.
\newblock In \emph{Proc. CVPR}, 2017.

\bibitem[Jiang et~al.(2018)Jiang, Sun, Jampani, Yang, Learned-Miller, and
  Kautz]{superslomo}
H.~Jiang, D.~Sun, V.~Jampani, M.-H. Yang, E.~Learned-Miller, and J.~Kautz.
\newblock Super slomo: High quality estimation of multiple intermediate frames
  for video interpolation.
\newblock 2018.

\bibitem[Jin et~al.(2011)Jin, Goto, and Ngan]{jin2011composite}
X.~Jin, S.~Goto, and K.~N. Ngan.
\newblock Composite model-based dc dithering for suppressing contour artifacts
  in decompressed video.
\newblock \emph{IEEE Transactions on Image Processing}, 20\penalty0
  (8):\penalty0 2110--2121, 2011.

\bibitem[Krizhevsky et~al.(2012)Krizhevsky, Sutskever, and
  Hinton]{Krizhevsky-et-al-NIPS12}
A.~Krizhevsky, I.~Sutskever, and G.~Hinton.
\newblock {ImageNet} classification with deep convolutional neural networks.
\newblock In \emph{NIPS}, 2012.

\bibitem[LeCun et~al.(1998)LeCun, Bottou, Bengio, and
  Haffner]{LeCun-et-al-IEEE98}
Y.~LeCun, L.~Bottou, Y.~Bengio, and P.~Haffner.
\newblock Gradient-based learning applied to document recognition.
\newblock \emph{Proceedings of the IEEE}, 86\penalty0 (11):\penalty0
  2278--2324, 1998.

\bibitem[Lee et~al.(2006)Lee, Lim, Park, Kim, and Ahn]{lee2006directcontrast}
J.~W. Lee, B.~R. Lim, R.-H. Park, J.-S. Kim, and W.~Ahn.
\newblock Two-stage false contour detection using directional contrast and its
  application to adaptive false contour reduction.
\newblock \emph{IEEE Transactions on Consumer Electronics}, 52\penalty0
  (1):\penalty0 179--188, 2006.

\bibitem[Liu et~al.(2017{\natexlab{a}})Liu, Sun, and Liu]{liu2017bitCNN}
J.~Liu, W.~Sun, and Y.~Liu.
\newblock Bit-depth enhancement via convolutional neural network.
\newblock In \emph{International Forum on Digital TV and Wireless Multimedia
  Communications}, pages 255--264. Springer, 2017{\natexlab{a}}.

\bibitem[Liu et~al.(2017{\natexlab{b}})Liu, Yeh, Tang, Liu, and
  Agarwala]{Liu-etal-ICCV17}
Z.~Liu, R.~A. Yeh, X.~Tang, Y.~Liu, and A.~Agarwala.
\newblock Video frame synthesis using deep voxel flow.
\newblock In \emph{Proc. ICCV}, 2017{\natexlab{b}}.

\bibitem[Long et~al.(2016)Long, Kneip, Alvarez, Li, Zhang, and
  Yu]{long2016matching}
G.~Long, L.~Kneip, J.~M. Alvarez, H.~Li, X.~Zhang, and Q.~Yu.
\newblock Learning image matching by simply watching video.
\newblock In \emph{Proc. ECCV}, pages 434--450. Springer, 2016.

\bibitem[Lucas and Kanade(1981)]{Lucas-Kanade-IUW81}
B.~Lucas and T.~Kanade.
\newblock An iterative image registration technique with an application to
  stereo vision.
\newblock In \emph{Proceedings of Imaging Understanding Workshop}, 1981.

\bibitem[Natsume et~al.(2018)Natsume, Saito, Huang, Chen, Ma, Li, and
  Morishima]{natsume2018siclope}
R.~Natsume, S.~Saito, Z.~Huang, W.~Chen, C.~Ma, H.~Li, and S.~Morishima.
\newblock Siclope: Silhouette-based clothed people.
\newblock \emph{arXiv preprint arXiv:1901.00049}, 2018.

\bibitem[Niklaus et~al.(2017{\natexlab{a}})Niklaus, Mai, and
  Liu]{Niklaus_CVPR_2017}
S.~Niklaus, L.~Mai, and F.~Liu.
\newblock Video frame interpolation via adaptive convolution.
\newblock In \emph{Proc. CVPR}, 2017{\natexlab{a}}.

\bibitem[Niklaus et~al.(2017{\natexlab{b}})Niklaus, Mai, and
  Liu]{Niklaus_ICCV_2017}
S.~Niklaus, L.~Mai, and F.~Liu.
\newblock Video frame interpolation via adaptive separable convolution.
\newblock In \emph{Proc. ICCV}, 2017{\natexlab{b}}.

\bibitem[Radford et~al.(2015)Radford, Metz, and Chintala]{Radford-etal-arXiv15}
A.~Radford, L.~Metz, and S.~Chintala.
\newblock Unsupervised representation learning with deep convolutional
  generative adversarial networks.
\newblock arXiv:1511.06434, 2015.

\bibitem[Ren et~al.(2016)Ren, Hu, Tai, Wang, Xu, Sun, and Yan]{ren2016look}
J.~S. Ren, Y.~Hu, Y.-W. Tai, C.~Wang, L.~Xu, W.~Sun, and Q.~Yan.
\newblock Look, listen and learn - a multimodal lstm for speaker
  identification.
\newblock In \emph{Proceedings of the 30th AAAI Conference on Artificial
  Intelligence}, pages 3581--3587, 2016.

\bibitem[Ronneberger et~al.(2015)Ronneberger, Fischer, and
  Brox]{Ronneberger-etal-MICAI15}
O.~Ronneberger, P.~Fischer, and T.~Brox.
\newblock U-net: Convolutional networks for biomedical image segmentation.
\newblock In \emph{Medical Image Computing and Computer-Assisted Intervention},
  2015.

\bibitem[R{\"o}ssler et~al.(2018)R{\"o}ssler, Cozzolino, Verdoliva, Riess,
  Thies, and Nie{\ss}ner]{rossler2018faceforensics}
A.~R{\"o}ssler, D.~Cozzolino, L.~Verdoliva, C.~Riess, J.~Thies, and
  M.~Nie{\ss}ner.
\newblock Faceforensics: A large-scale video dataset for forgery detection in
  human faces.
\newblock \emph{arXiv preprint arXiv:1803.09179}, 2018.

\bibitem[Simonyan and Zisserman(2014)]{Simonyan-Zisserman-arXiv14}
K.~Simonyan and A.~Zisserman.
\newblock Very deep convolutional networks for large-scale image recognition.
\newblock arXiv:1409.1556, 2014.

\bibitem[Tai et~al.(2017)Tai, Yang, and Liu]{Tai-DRRN-2017}
Y.~Tai, J.~Yang, and X.~Liu.
\newblock Image super-resolution via deep recursive residual network.
\newblock In \emph{Proc. CVPR}, 2017.

\bibitem[Wan et~al.(2016)Wan, Cheung, Florencio, Zhang, and Au]{wan2016bitMAP}
P.~Wan, G.~Cheung, D.~Florencio, C.~Zhang, and O.~C. Au.
\newblock Image bit-depth enhancement via maximum a posteriori estimation of ac
  signal.
\newblock \emph{IEEE Transactions on Image Processing}, 25\penalty0
  (6):\penalty0 2896--2909, 2016.

\bibitem[Wang et~al.(2014)Wang, Guo, Zhu, Wang, and Wang]{wang2014video}
C.~Wang, Y.~Guo, J.~Zhu, L.~Wang, and W.~Wang.
\newblock Video object co-segmentation via subspace clustering and quadratic
  pseudo-boolean optimization in an mrf framework.
\newblock \emph{IEEE Transactions on Multimedia}, 16\penalty0 (4):\penalty0
  903--916, 2014.

\bibitem[Wang et~al.(2017)Wang, Zhu, Guo, and Wang]{wang2017video}
C.~Wang, J.~Zhu, Y.~Guo, and W.~Wang.
\newblock Video vectorization via tetrahedral remeshing.
\newblock \emph{IEEE Transactions on Image Processing}, 26\penalty0
  (4):\penalty0 1833--1844, 2017.

\bibitem[Wang et~al.(2018)Wang, Huang, Han, and Wang]{wang2018videoinp}
C.~Wang, H.~Huang, X.~Han, and J.~Wang.
\newblock Video inpainting by jointly learning temporal structure and spatial
  details.
\newblock \emph{arXiv preprint arXiv:1806.08482}, 2018.

\bibitem[Yoo et~al.(2009)Yoo, Song, and Sohn]{yoo2009loop}
K.~Yoo, H.~Song, and K.~Sohn.
\newblock In-loop selective processing for contour artefact reduction in video
  coding.
\newblock \emph{Electronics letters}, 45\penalty0 (20):\penalty0 1020--1022,
  2009.

\bibitem[Yu and Koltun(2016)]{yu2015multi}
F.~Yu and V.~Koltun.
\newblock Multi-scale context aggregation by dilated convolutions.
\newblock In \emph{ICLR}, 2016.

\bibitem[Zhao et~al.(2018)Zhao, Chen, Xing, Li, Bessinger, Liu, Zuo, and
  Yang]{zhao2018identity}
Y.~Zhao, W.~Chen, J.~Xing, X.~Li, Z.~Bessinger, F.~Liu, W.~Zuo, and R.~Yang.
\newblock Identity preserving face completion for large ocular region
  occlusion.
\newblock \emph{arXiv preprint arXiv:1807.08772}, 2018.

\end{thebibliography}
}

\end{document}